%% file: arxiv.tex
\documentclass[runningheads]{llncs}

\usepackage[mobile]{eccv}

\usepackage{eccvabbrv}
\usepackage{makecell} 
\usepackage{tabularx} 
\usepackage{graphicx}
\usepackage{subcaption} 
\usepackage{booktabs}
\usepackage{multirow}
\newcommand{\mhan}[1]{\textcolor{red}{#1}}
\newcommand{\blues}[1]{\textcolor{blue}{#1}}
\usepackage[accsupp]{axessibility}  

\usepackage{hyperref}
\usepackage{pgfplots}
\pgfplotsset{compat=1.17} 
\usepackage{orcidlink}
\usepackage{fvextra}
\usepackage[most]{tcolorbox}

\begin{document}

\begingroup
\renewcommand{\thefootnote}{*}
\footnotetext{Equal contribution.}
\endgroup

\title{Beyond Dense Futures: World Models as Structured Planners for Robotic Manipulation} 

\titlerunning{World Models as Structured Planners for Robotic Manipulation}

\author{Minghao Jin\inst{1}$^{*}$, Mozheng Liao\inst{1}$^{*}$, Mingfei Han\inst{1,2}$^{*}$, Zhihui Li\inst{1}, Xiaojun Chang\inst{1,2}}

\authorrunning{M. Jin et al.}

\institute{University of Science and Technology of China \and
Department of CV, MBZUAI \\
\url{https://wm-planner.github.io/structvla/} }

\maketitle

\input{sec/0_abstract}
\input{sec/1_intro}
\input{sec/2_related_work}
\input{sec/3_method}

\input{sec/4_experiments}

\input{sec/5_conclusion}


%
%
\bibliographystyle{splncs04}
\bibliography{main}
\newpage
\appendix
\input{sec/6_supplementation}
\end{document}

%% file: sec/0_abstract.tex

\begin{abstract}
Recent world-model-based Vision-Language-Action (VLA) architectures have improved robotic manipulation through predictive visual foresight. However, dense future prediction introduces visual redundancy and accumulates errors, causing long-horizon plan drift. Meanwhile, recent sparse methods typically represent visual foresight using high-level semantic subtasks or implicit latent states. These representations often lack explicit kinematic grounding, weakening the alignment between planning and low-level execution. To address this, we propose StructVLA, which reformulates a generative world model into an explicit structured planner for reliable control. Instead of dense rollouts or semantic goals, StructVLA predicts sparse, physically meaningful structured frames. Derived from intrinsic kinematic cues (\eg, gripper transitions and kinematic turning points), these frames capture spatiotemporal milestones closely aligned with task progress. We implement this approach through a two-stage training paradigm with a unified discrete token vocabulary: the world model is first trained to predict structured frames and subsequently optimized to map the structured foresight into low-level actions. This approach provides clear physical guidance and bridges visual planning and motion control. In our experiments, StructVLA achieves strong average success rates of 75.0\% on SimplerEnv-WidowX and 94.8\% on LIBERO. Real-world deployments further demonstrate reliable task completion and robust generalization across both basic pick-and-place and complex long-horizon tasks.
  \keywords{Vision language action model \and  World model \and Robotic manipulation}
\end{abstract}

%% file: sec/1_intro.tex
\section{Introduction}
\label{sec:intro}
\begin{figure}[t]
  \centering
  \includegraphics[width=\columnwidth]{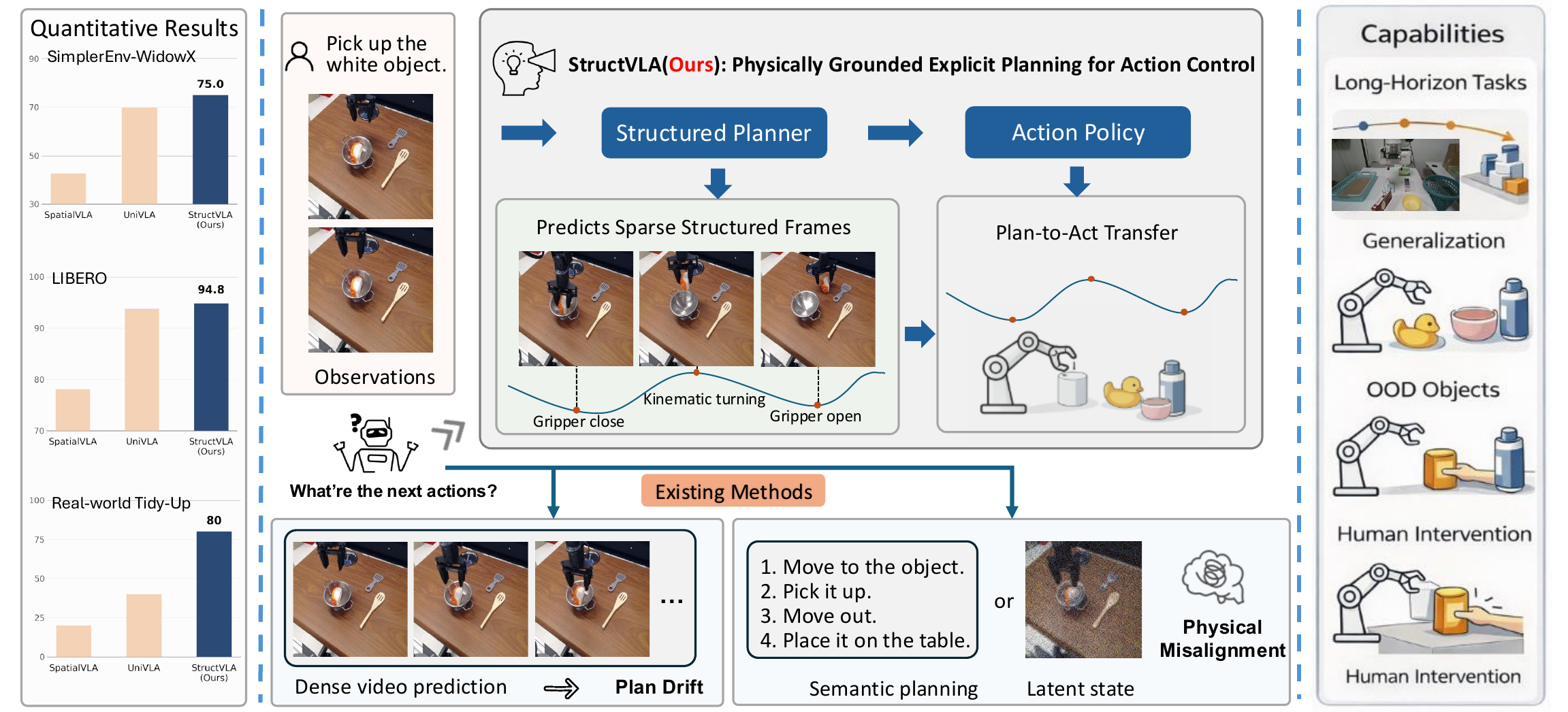}
  \caption{{\bf Illustration of StructVLA (Structured Planner for Vision-Language-Action).} Existing generative VLA methods either predict dense future observations \textit{(bottom left)}, which accumulate errors and cause long-horizon plan drift, or rely on semantic or latent planning \textit{(bottom right)} that lacks explicit geometric grounding and leads to physical misalignment. Our StructVLA \textit{(top)} learns a physically grounded structured planner by training a world model to predict sparse structured frames, and transfers this representation to action control. This design enables stable long-horizon manipulation and demonstrates strong generalization and robustness across simulation and real-world benchmarks.}

  \label{fig:intro}
  \vspace{-3mm}
\end{figure}

Robotic manipulation typically requires coupling coherent task planning with reliable physical control. Vision-Language-Action (VLA) models \cite{black2024pi0visionlanguageactionflowmodel,brohan2022rt,zitkovich2023rt,kim2024openvla,chen2025villa,feng2025generalist,li2024cogact,qu2025spatialvla,shenvideovla,sun2025geovla,zhong2025flowvla,zhang2025dreamvla,wang2025unifiedvisionlanguageactionmodel,bu2025univla,hu2026bagelvla,kim2026cosmos} address this challenge by grounding natural language instructions in visual observations to directly output low-level actions. Recently, the integration of generative world models into VLA architectures \cite{wang2025unifiedvisionlanguageactionmodel,gao2025adaworld, tharwat2025latent} has emerged as a powerful paradigm for long-horizon manipulation. By internalizing environmental dynamics, these generative architectures provide predictive visual foresight, offering spatiotemporal context that enables the policy to anticipate future states beyond instantaneous perception.

Despite these advantages, effectively representing this visual foresight remains a fundamental challenge. Many approaches represent foresight as dense or step-wise future predictions (pixel or latent) \cite{wang2025unifiedvisionlanguageactionmodel, shenvideovla, chen2025large,kim2026cosmos,zhang2025dreamvla}, which can amplify long-horizon errors when used as direct guidance. Moreover, dense representations contain massive visual redundancy, often obscuring essential task events within high-frequency, task-irrelevant noise. In addition, generating continuous temporal rollouts tend to accumulate prediction errors. Over extended horizons, local visual artifacts or physical inconsistencies compound and can result in plan drift.

To mitigate the visual redundancy and error accumulation from dense rollouts, recent methods have explored sparse visual foresight \cite{zhang2026foreact,long2026scaling,hu2026bagelvla}. While these paradigms effectively compress the prediction horizon, existing sparse frameworks also introduce structural limitations. Some approaches define subgoals in semantic space, where high-level abstractions may not specify the precise geometric grounding required for manipulation. Other approaches rely on implicit latent representations \cite{xie2025latent,zhang2025dreamvla}; while these methods facilitate policy learning, they lack explicit, verifiable geometric targets for closed-loop control. Without clear physical grounding, both representations struggle to provide actionable guidance for low-level kinematic execution, motivating a planning representation that is both physically grounded and control-effective.

To address these fundamental limitations, we propose \textbf{StructVLA}, a framework that reformulates the generative world model into an explicit, physically grounded structured planner. StructVLA predicts sparse structured frames as mid-level visual subgoals that mark essential manipulation milestones. Rather than using semantic subtask boundaries to define subgoals, we derive these frames from intrinsic kinematic cues, such as gripper state transitions  and kinematic turning points. Gripper transitions capture contact-state changes, while kinematic turning points reflect meaningful shifts in motion; together they provide progress anchors that are mechanically meaningful across tasks.  As a result, the generated visual foresight provides actionable spatiotemporal cues and is tightly coupled to executable interaction phases. 

As illustrated in \cref{fig:overview}, StructVLA employs a two-stage training paradigm. In the first stage, the model autoregressively predicts explicit, discrete structured frames, which serve as explicit geometric targets and help reduce long-horizon drift. In the second stage, the model is optimized to generate low-level action tokens, leveraging the structured-planning capability learned in Stage 1. By representing vision, language, and action in a shared discrete vocabulary, StructVLA strengthens the coupling between visual planning and motion control.

Our main contributions are summarized as follows:
\begin{itemize}
    \item \textbf{Physics-grounded structured planning targets.} We introduce an automated data curation pipeline that extracts structured frames from robot-centric kinematic cues, providing sparse, mechanically meaningful anchors without semantic subtask labeling.
    \item \textbf{Unified StructVLA framework.} We present StructVLA, a unified framework that reformulates a world model into an explicit structured planner for control. It is trained in two stages: first autoregressively predicting structured frames as mid-level visual subgoals, then generating low-level action tokens by leveraging the structured-planning capability learned in the first stage, within a shared discrete token space.
    \item \textbf{Strong empirical results.} StructVLA achieves average success rates of 75.0\% on SimplerEnv-WidowX \cite{li2024evaluating} and 94.8\% on LIBERO \cite{liu2023libero}, and demonstrates reliable task completion  and robust generalization in real-world evaluations on both basic pick-and-place and complex long-horizon tasks.
\end{itemize}

\vspace{-4mm}

%% file: sec/2_related_work.tex
\section{Related Work}
\label{sec:related_work}

\textbf{Vision-Language-Action Models.} Recent vision-language-action (VLA) models \cite{black2024pi0visionlanguageactionflowmodel,brohan2022rt,zitkovich2023rt,kim2024openvla,chen2025villa,feng2025generalist,li2024cogact,qu2025spatialvla,shenvideovla,sun2025geovla,zhong2025flowvla,zhang2025dreamvla,wang2025unifiedvisionlanguageactionmodel,bu2025univla,hu2026bagelvla,kim2026cosmos} have demonstrated strong performance for robotic manipulation. A common line of work builds on large vision-language models (VLMs) and learns to map multimodal representations to robot actions. For instance, OpenVLA \cite{kim2024openvla} directly predicts action tokens from VLM, while $\pi_0$ \cite{black2024pi0visionlanguageactionflowmodel} and CogACT \cite{li2024cogact} augment the backbone with a flow-matching or diffusion-based action expert to improve continuous control. Villa-X \cite{chen2025villa} further introduces an intermediate latent-action expert before generating final actions. Recent extensions also incorporate 3D cues to strengthen spatial grounding, such as SpatialVLA \cite{qu2025spatialvla} and GeoVLA \cite{sun2025geovla}. Another line of work explores generative action modeling and video-diffusion-based control. Diffusion Policy \cite{chi2023diffusion} formulates action generation directly as a diffusion process, while VPP \cite{hu2024video} exploits representations from video diffusion models for control. VIDAR \cite{feng2025generalist} integrates action reasoning into a video diffusion model, and VideoVLA \cite{shenvideovla} further studies generalization in video-diffusion-based VLA frameworks. FlowVLA \cite{zhong2025flowvla} uses optical flow as a motion-foresight cue for action prediction. DreamVLA \cite{zhang2025dreamvla} uses world-model forecasting to guide long-horizon action generation. In contrast, our planner predicts kinematically grounded structured frames as explicit progress anchors, making foresight directly actionable for long-horizon manipulation.

\noindent\textbf{Generative models for robotic manipulation.} Many studies have applied world models \cite{Ha2018worldmodel,hafner2020dreamcontrollearningbehaviors,lecun2022pathtomachine, xu2022cascade} to robotic manipulation across both simulation and real-robot settings. DayDreamer \cite{pmlr-v205-wu23c} introduces the Dreamer algorithm into real-world robot learning, improving sample efficiency. CASCADE \cite{xu2022cascade} expands world-model training coverage via collaborative exploration to improve generalization. SWIM \cite{mendonca2023structured} trains world models on human demonstration videos and uses them to guide manipulation. UWM \cite{zhu2025uwm} further unifies video diffusion and action diffusion for joint modeling of dynamics and policy learning. UniVLA \cite{wang2025unifiedvisionlanguageactionmodel} unifies vision, language, and action as an autoregressive discrete sequence and incorporates world-model post-training to capture temporal and causal structure for downstream policy learning. For joint forecasting, VideoVLA \cite{shenvideovla} employs a multimodal DiT \cite{peebles2023scalable} to jointly forecast an action sequence and future visual outcomes. Large Video Planner \cite{chen2025large} leverages foundation-scale video pretraining to generate dense video plans, which are then post-processed into executable robot actions. Similarly, Cosmos Policy \cite{kim2026cosmos} adapts a pretrained video diffusion model via latent frame injection, and can optionally perform value-guided best-of-$N$ planning to evaluate candidate action trajectories, incurring higher test-time sampling cost.
In contrast, we explore world models as explicit sparse structured planners, producing physically grounded progress anchors that are directly actionable for closed-loop control.

\noindent\textbf{Task planning in robot learning.} Task planning is crucial for long-horizon manipulation. Many studies have explored decomposing language instructions into sub-tasks, such as utilizing code-generation models \cite{codeaspolicies2022} to translate natural language into executable policies. Text2Motion \cite{lin2023text2motion} converts both long and short natural language instructions into symbolic task goals and feasible action sequences, while ThinkAct \cite{huang2025thinkact} leverages VLM reasoning for trajectory-level planning. Complementary to language-centric planning, other approaches derive planning signals from visual dynamics. Track2Act \cite{bharadhwaj2024track2act} predicts pixel-level trajectories from internet videos to assist manipulation, and LDP \cite{xie2025latent} models latent actions between adjacent frames for trajectory planning. UniVLA (Bu et al.) \cite{bu2025univla} extracts task-centric latent actions from unannotated videos to enable cross-embodiment policy learning. More recently, generative foresight has been used to instantiate planners in different forms. ForeAct \cite{zhang2026foreact} and VISTA \cite{long2026scaling} employ generative models as high-level planners that provide subgoal guidance to separate low-level VLAs, while BagelVLA \cite{hu2026bagelvla} and mimic-video \cite{pai2025mimic} incorporate planning signals through implicit latent foresight. In particular, BagelVLA \cite{hu2026bagelvla} and ForeAct \cite{zhang2026foreact} adopt subtask-anchored foresight, where subgoals are specified or sampled within semantic subtask segments. In contrast, StructVLA derives structured subgoals from kinematic interaction cues and reformulates the world model as a kinematically grounded planner, providing control-effective visual anchors for action generation.


%% file: sec/3_method.tex
\newcommand{\placeholder}[2]{%
  \fbox{\parbox[c][#2][c]{#1}{\centering \vspace{0.5em}\textsf{Placeholder}\vspace{0.5em}}}%
}

\section{Method}
\label{sec:method}
\begin{figure*}[t]
  \centering
  \includegraphics[width=1.0\textwidth]{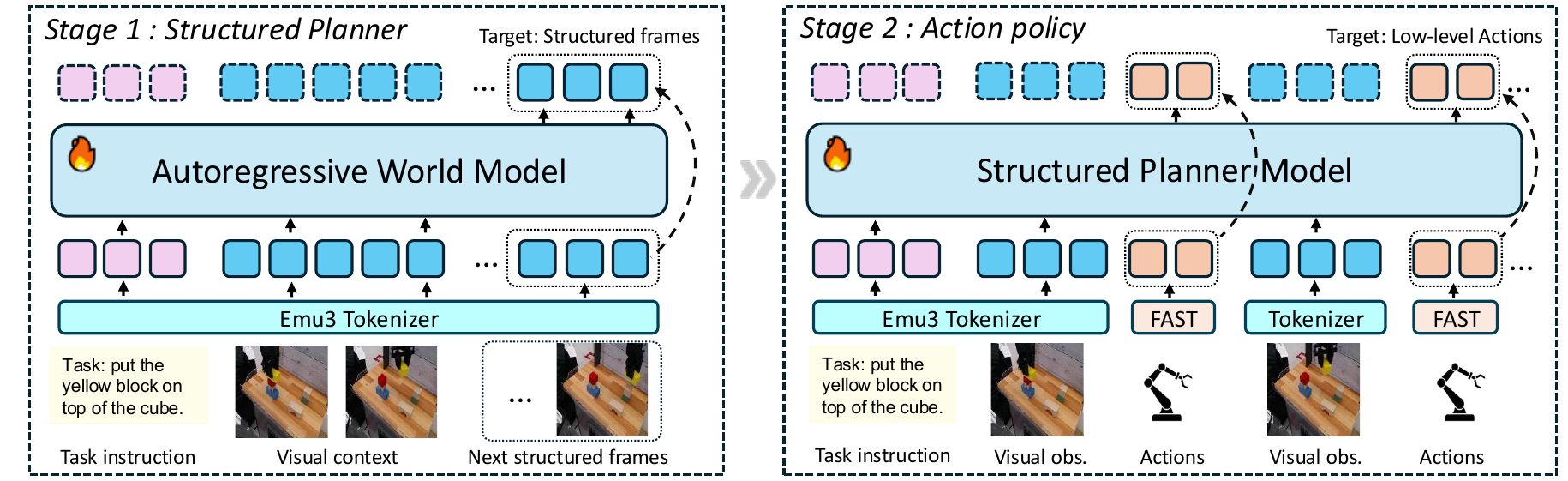}

  \caption{\textbf{Overview of StructVLA.}
StructVLA is trained in two stages.
\textbf{Stage 1 (Structured Planner):} an autoregressive world model predicts sparse structured frames that capture physically grounded progress anchors, conditioned on the instruction and visual context. \textbf{Stage 2 (Action Policy):} we fine-tune the structured planner for control by conditioning on the instruction together with interleaved observation and actions, transferring structured planning into low-level control.}

  \label{fig:overview}
  \vspace{-5mm}
\end{figure*}

\cref{fig:overview} summarizes the overall design of StructVLA. We reformulate a single generative world model as a structured planner by introducing sparse structured frames as physically grounded visual subgoals. \cref{sec:Data construction} describes an automated data curation pipeline for extracting these structured frames. We train StructVLA with a two-stage paradigm: \cref{sec: train structured planner} details Stage 1 for structured frame prediction, and \cref{sec: action policy} presents Stage 2 for transferring the structured foresight to action policy learning.

\subsection{Physics-Grounded Data Curation}\label{sec:Data construction}

At the core of our structured planner are structured frames, which serve as physically grounded visual subgoals. During offline data curation, an automated pipeline extracts these frames from raw robotic trajectories to exclusively supervise the generative world model. Instead of representing foresight as dense continuous rollouts, structured frames provide compact progress anchors that connect task intent to executable motion phases. By deriving them from robot-centric kinematic cues, our curation produces subgoals that are consistently aligned with interaction transitions in manipulation.

Formally, consider a robotic manipulation trajectory $\mathcal{T}=\big(l,\{(o_t,a_t)\}_{t=0}^T\big)$, where $l$ denotes the language instruction, $o_t$ is the visual observation, and $a_t$ represents the executed action at timestep $t$. Within a trajectory, the most intuitive anchors are explicit milestones (\eg, object grasped or placed), which define clear subtask boundaries. However, relying only on these sparse events omits critical task-progression details. In practice, many informative intermediate states are implicitly reflected in the motion trajectory, such as the kinematic turning point where the gripper aligns above a target before shifting direction to descend. To capture such physically meaningful states without manual annotation, we introduce an automated extraction procedure based on kinematic cues. Our offline pipeline first proposes candidate frames from robot-centric kinematic cues, and then applies filtering to remove redundancy and refine the candidates.

\noindent{\bf Kinematic Structured Frame Extraction.}
To ensure robustness across diverse manipulation speeds and tasks, we compute adaptive kinematic thresholds for each episode from the distribution of end-effector motion magnitudes $\|v_t\|$, which is derived from the actions $a_t$. Combined with Exponential Moving Average (EMA) smoothing to filter out high-frequency control noise, we extract structured frames using two primary kinematic cues:

\noindent\textit{Gripper State Transitions:} We extract structured frames around gripper state changes (open or close), which often mark contact state transitions. Since raw gripper signals can be noisy and the physical actuation may lag behind the signal flip in practice, we apply a stabilization window $\delta_{\text{settle}}$ after each flip time $t_c$. We then select the last quasi static observation $o_{t^*}$ immediately before the robot resumes a major motion, identified when the smoothed speed exceeds an adaptive threshold $\tau_{\text{high}}$. Formally, for each flip time $t_c$, we define
\begin{equation}
    S_{\text{grip}} = \Big\{ o_{t^*} \;\Big|\; t^* = \min \big\{ t > t_c + \delta_{\text{settle}} \mid \text{EMA}(\|v_t\|) > \tau_{\text{high}} \big\} - 1 \Big\}
\end{equation}

\noindent\textit{Kinematic Turning Points:} Even with a fixed gripper state, manipulation often involves brief slowdowns during motion regime transitions. For example, the end-effector may complete lateral alignment above the target and stabilize before initiating the downward approach. We identify such moments by applying a sliding window of length $W$ and selecting frames whose smoothed speed remains below an adaptive threshold $\tau_{\text{low}}$ throughout the window ending at timestep $t$:
\begin{equation}
S_{\text{turn}} =
\big\{ o_t \;\big|\;
\max_{k \in [t-W+1, t]} \mathrm{EMA}\!\left(\|v_k\|\right)
\le \tau_{\text{low}}
\big\}
\end{equation}

\noindent \textit{Temporal Coverage:} To avoid extracting redundant frames from a prolonged low-speed segment, we enforce a minimum temporal gap between selected frames. For unusually long segments where distinct kinematic events are sparse, we apply a gap filling step to ensure adequate subgoal coverage. Specifically, we sample additional observations $S_{\text{fill}}$ by selecting frames with the smallest $L_2$ change in consecutive action commands, $\Delta a_t=\|a_t-a_{t-1}\|_2$, which serves as a proxy for low-motion, stable states within the segment. The union of these sets forms our initial candidate structured frame set $\mathcal{S} = S_{\text{grip}} \cup S_{\text{turn}} \cup S_{\text{fill}}$.

\noindent{\bf Offline Trajectory Filtering.} While kinematic cues reliably detect interaction transitions, they provide limited task-level context and may occasionally propose false positives under hardware perturbations. To refine the final structured frames, we use a VLM, \ie, GPT-5.2~\cite{openai2025gpt52}, strictly for offline refinement. Specifically, we provide the global instruction $l$ together with the candidate observations $\mathcal{S}$, and use a fixed prompt (detailed in the Appendix) with filtering rules to remove visually redundant or task-irrelevant frames. This verification refines the preliminary structured frames obtained from kinematic cues, so that the final sequence better reflects coherent and executable milestones. Crucially, the filtering is applied only offline during dataset construction. We train StructVLA solely on the curated data. During inference, we deploy the action policy with no additional curation or VLM assistance.

\subsection{Structured Planner (Stage 1: Visual Foresight)}\label{sec: train structured planner}

To achieve this structured planning, we train the generative world model to predict the physically grounded structured frames $s \in \mathcal{S}$ identified during data curation. Formally, given a language instruction $l$ and a visual context window $\mathcal{O}_t = \{ o_{t-H+1}, \dots , o_t\}$, the model learns the conditional distribution $P_\theta(s \mid l, \mathcal{O}_t)$ of a future structured frame $s$.

We employ the Emu3 \cite{wang2024emu3} tokenizer with Vector Quantization (VQ) to discretize the language instructions and image frames into a shared vocabulary. Specifically, the language instruction $l$ is tokenized into $T_l$, while the visual context window $\mathcal{O}_t$ is transformed into a sequence of tokens $T_{\mathcal{O}_t} = \{T_{o_{t-H+1}}, \dots, T_{o_t}\}$, where each $T_{o_i}$ represents the discretized features of observation $o_i$. Similarly, the target structured frame $s$ is represented by a discrete token sequence $T_s = \{z_1, z_2, \dots, z_K\}$.

Following the standard autoregressive generation paradigm, the model is trained to minimize the cross-entropy loss over the predicted visual tokens $T_s$, conditioned on the tokenized instruction and historical observations:
\begin{equation}
    \mathcal{L}_{\text{planner}} = - \sum_{k=1}^{K} \log P_\theta(z_k \mid z_{<k}, T_l, T_{\mathcal{O}_t})
\end{equation}

Through sequential token prediction, the model learns to forecast kinematically grounded structured frames, capturing key transitions in manipulation and developing an implicit awareness of task progress in its visual foresight. This provides stable visual guidance for subsequent action control.


\subsection{Action Policy (Stage 2: Plan-to-Act Transfer)}\label{sec: action policy}

In Stage 2, the pretrained structured planner is directly formulated as an executable action policy by shifting the optimization objective to predicting action tokens. We discretize continuous actions into discrete tokens using the FAST tokenizer \cite{pertsch2025fast}. Crucially, these action tokens are integrated into a unified vocabulary shared with the visual and language modalities, enabling the model to directly output control signals through standard autoregressive generation.

During training, the conditioning context is explicitly constructed by concatenating the tokenized global language instruction $T_l$ with interleaved visual and action tokens. Specifically, for a given timestep $t$ and a context window of length $H$, this multimodal history sequence is defined as $\mathcal{H}_t = \{T_{o_{t-H+1}}, T_{a_{t-H+1}}, \dots, T_{o_t}\}$, where $T_{o_i}$ and $T_{a_i}$ represent the discretized tokens of the observation and action at timestep $i$, respectively. Given this explicitly interleaved sequence, the model is optimized to autoregressively predict the sequence of discretized action tokens $T_{a_t} = \{a_1, a_2, \dots, a_M\}$ required to execute the current subtask.

Following the standard next-token prediction paradigm, we restrict the supervision signal strictly to the action domain. The cross-entropy loss is computed exclusively over these sequentially generated action tokens:
\begin{equation}
\mathcal{L}_{\text{act}} = - \sum_{m=1}^{M} \log P_\theta(a_m \mid a_{<m}, T_l, \mathcal{H}_t)
\end{equation}
By dynamically shifting the prediction target within a unified vocabulary, this formulation allows the generative structured planner to directly translate its internalized visual foresight into reliable, low-level action control.

%% file: sec/4_experiments.tex
\begin{table}[t]
  \centering
  
  \begin{minipage}[t]{0.48\linewidth}
    \centering
    \caption{\textbf{Results on SimplerEnv \cite{li2024evaluating}.} Each model is evaluated for task success rate across four manipulation tasks. StructVLA yields the strongest average performance.}
    \label{tab:simpler_env_widowx}
    \vspace{2mm} 
    \resizebox{\linewidth}{!}{%
      \begin{tabular}{@{} l ccccc @{}}
        \toprule
        \textbf{Model} & Spoon & Carrot & Stack & Eggplant & \textbf{Avg.} \\
        \midrule
        RT-1-X \cite{o2024open} & 0.0\% & 4.2\% & 0.0\% & 0.0\% & 1.1\% \\
        Octo-Base \cite{team2024octo} & 12.5\% & 31.9\% & 0.0\% & 43.1\% & 16.0\% \\
        Octo-Small \cite{team2024octo} & 47.2\% & 9.7\% & 4.2\% & 56.9\% & 29.5\% \\
        OpenVLA \cite{kim2024openvla} & 0.0\% & 0.0\% & 0.0\% & 4.1\% & 1.0\% \\
        RoboVLMs \cite{liu2025towards} & 45.8\% & 20.8\% & 4.2\% & 79.2\% & 37.5\% \\
        SpatialVLA \cite{qu2025spatialvla} & 16.7\% & 25.0\% & 29.2\% & \textbf{100\%} & 42.7\% \\
        VideoVLA \cite{shenvideovla} & 75.0\% & 20.8\% & \textbf{45.8\%} & 70.8\% & 53.1\% \\
        mimic-video \cite{pai2025mimic} & 41.7\% & 54.2\% & 29.2\% & \textbf{100\%} & 56.3\% \\
        UniVLA \cite{wang2025unifiedvisionlanguageactionmodel} & 83.3\% & 66.7\% & 33.3\% & 95.8\% & 69.8\% \\
        \midrule
        \textbf{StructVLA (Ours)} & \textbf{87.5\%} & \textbf{75.0\%} & \textbf{45.8\%} & 91.7\% & \textbf{75.0\%} \\
        \bottomrule
      \end{tabular}%
    }
  \end{minipage}
  \hfill 
  \begin{minipage}[t]{0.48\linewidth}
    \centering
    \caption{\textbf{Results on LIBERO \cite{liu2023libero}.} Comparison of task success rates across four suites. StructVLA outperforms prior baselines, especially on the LIBERO-Long suite. * denotes reproduced results.}
    \label{tab:libero_comparison}
    \vspace{2mm}
    \renewcommand{\arraystretch}{1.06}
    \resizebox{\linewidth}{!}{%
      \begin{tabular}{@{} l ccccc @{}}
        \toprule
        \textbf{Method} & SPATIAL & OBJECT & GOAL & LONG & \textbf{Avg.} \\
        \midrule
        DP~\cite{chi2023diffusion}          & 78.3\% & 92.5\% & 68.3\% & 50.5\% & 72.4\% \\
        Octo~\cite{team2024octo}       & 78.9\% & 85.7\% & 84.6\% & 51.1\% & 75.1\% \\
        OpenVLA~\cite{kim2024openvla} & 84.9\% & 88.4\% & 79.2\% & 53.7\% & 76.5\% \\
        SpatialVLA~\cite{qu2025spatialvla} & 88.2\% & 89.9\% & 78.6\% & 55.5\% & 78.1\% \\
        CoT-VLA~\cite{zhao2025cot}  & 87.5\% & 91.6\% & 87.6\% & 69.0\% & 81.1\% \\
        $\pi_{0}$-FAST~\cite{pertsch2025fast} & 96.4\% & 96.8\% & 88.6\% & 60.2\% & 85.5\% \\
        FlowVLA~\cite{zhong2025flowvla} & 93.2\% & 95.0\% & 91.6\% & 72.6\% & 88.1\% \\
        DreamVLA~\cite{zhang2025dreamvla} & \textbf{97.5\%} & 94.0\% & 89.5\% & 89.5\% & 92.6\% \\
        UniVLA*~\cite{wang2025unifiedvisionlanguageactionmodel}       & 96.4\% & \textbf{98.0\%} & 90.8\% & 89.6\% & 93.7\% \\
        \midrule
        \textbf{StructVLA (Ours)}       & 95.4\% & \textbf{98.0\%} & \textbf{93.4\%} & \textbf{92.2\%} & \textbf{94.8\%} \\
        \bottomrule
      \end{tabular}%
    }
  \end{minipage}
  \vspace{-2mm}
\end{table}

\begin{figure}[t]
  \centering
  
  \begin{minipage}[b]{0.48\textwidth} 
    \centering
    \includegraphics[width=\linewidth]{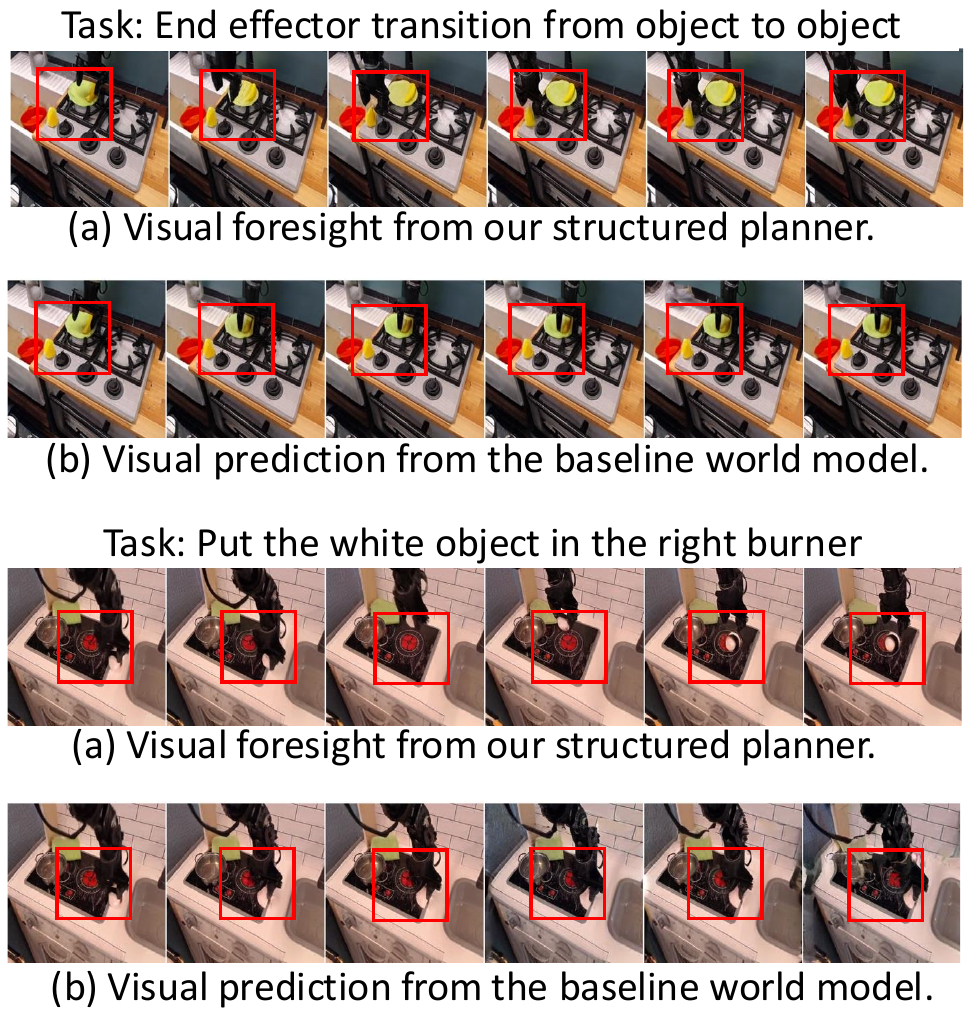}
    \caption{
        \textbf{Qualitative comparison of visual predictions.} Our structured planner generates coherent, long-horizon visual foresight, supporting robust planning, whereas the baseline world model struggles with long-range comprehension, producing short-horizon predictions with degraded image quality. Red boxes highlight key differences.
    }
    \label{fig:planner_visualization}
  \end{minipage}
  \hfill 
  \begin{minipage}[b]{0.48\textwidth} 
    \centering
    \includegraphics[width=\linewidth]{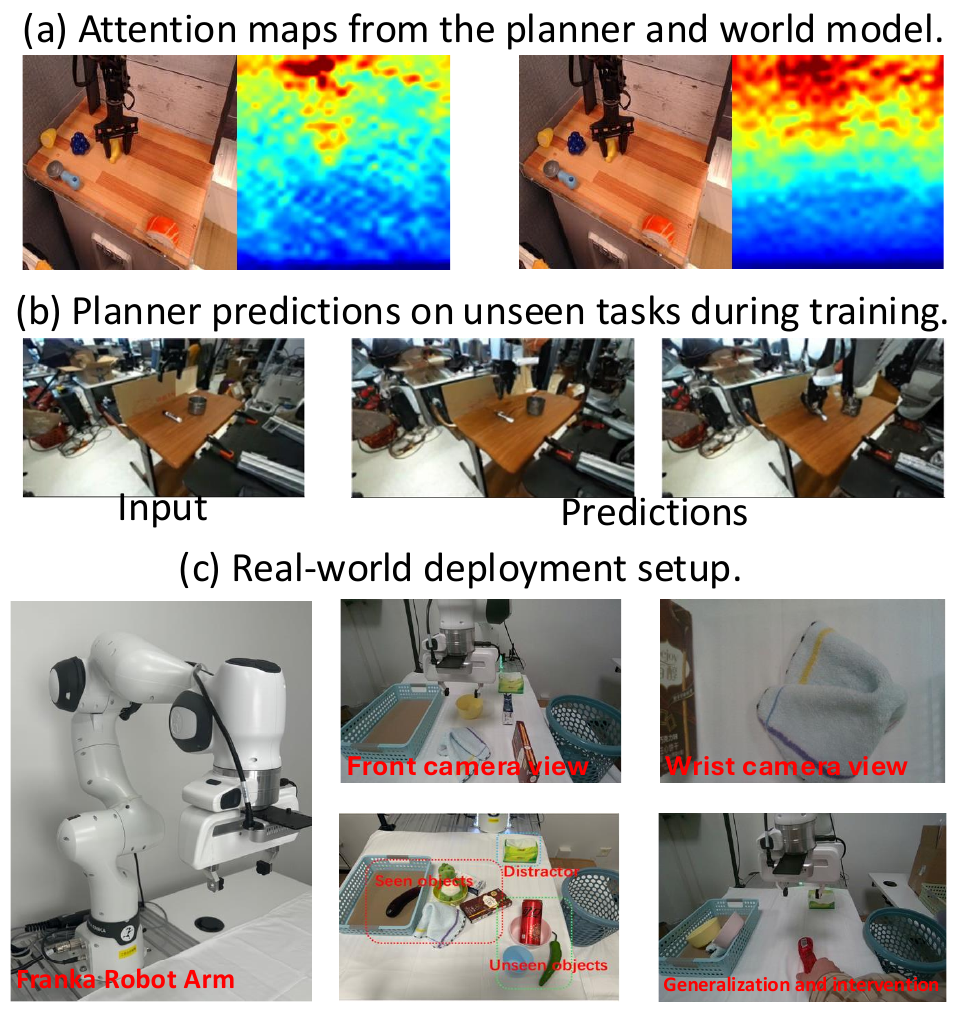}
    \caption{
    \textbf{More visualizations} (a) \textbf{Attention maps:} The planner (left) localizes task-critical interactions (\eg, gripper--object contact), while the baseline (right) is diffuse. (b) \textbf{OOD predictions:} StructVLA retains world-model priors, enabling zero-shot visual planning on unseen scenes and tasks. (c) \textbf{Real-world deployment setup.}
    }
    \label{fig:planner_more_visualization}
  \end{minipage}
  
  \vspace{-4mm} 
\end{figure}

\section{Experiments}
\label{sec:experiments}

In this section, we comprehensively evaluate StructVLA across simulation benchmarks and real-world deployments. \cref{sec:dataset} introduces the simulation and real-world deployment settings, together with the curated datasets used for training. \cref{sec:implementation} details the model architecture and training configurations. \cref{sec:experimental_evaluation} presents comparative results against representative baselines. Finally, \cref{sec: extend analysis} presents in-depth analyses and extended studies of the overall VLA pipeline.


\subsection{Experimental Setup}\label{sec:dataset}

\noindent\textbf{SimplerEnv benchmark.} 
SimplerEnv~\cite{li2024evaluating} evaluates transfer and generalization of policies trained on real-world robot trajectories under controlled distribution shifts (\eg, lighting, texture, and camera pose) across WidowX and Google Robot platforms. We mainly report results on WidowX, which is commonly used. We train StructVLA on the standard BridgeData V2~\cite{walke2023bridgedata} dataset.

\noindent\textbf{LIBERO benchmark.} The LIBERO benchmark~\cite{liu2023libero} evaluates multi-task manipulation and long-horizon consistency within a simulated Franka Emika Panda robot arm. There are four task suites including LIBERO-Spatial, LIBERO-Object, LIBERO-Goal, and LIBERO-Long (LIBERO-10). Specifically, each suite comprises 10 distinct tasks with 50 expert demonstrations per task, and we jointly train a single model across the combined datasets of all suites.

\noindent\textbf{Real-world evaluation.} 
We deploy StructVLA on a physical Franka arm to evaluate robustness under real-world noise and perturbations. Our setup includes four pick-and-place tasks (yellow bowl, biscuit box, toy, eggplant) and a long-horizon tidy-up task with distractors. We train a multi-task policy for the four tasks using 1,152 trajectories (288 per task) and a dedicated policy for tidy-up using 648 trajectories. We compare against UniVLA \cite{wang2025unifiedvisionlanguageactionmodel} under identical conditions and evaluate zero-shot generalization to unseen objects (novel colors and geometries) and robustness under human interventions.

\subsection{Implementation Details}\label{sec:implementation}

{\bf Model structure.} Our world model adopts a purely autoregressive Transformer architecture with 8.5B parameters, initialized from the world model of UniVLA\cite{wang2025unifiedvisionlanguageactionmodel}. Images are tokenized using the Emu3\cite{wang2024emu3} VQ-based encoder with a spatial compression factor of 8. For action tokenization, we first apply 1st and 99th percentile normalization, and then utilize the FAST\cite{pertsch2025fast} tokenizer with a vocabulary size of 1024. We integrate actions by mapping them to the final 1024 token IDs of the language vocabulary. Consequently, vision, language, and action share a single discrete token space, enabling joint representation learning and cross-modal alignment. Visual observation setups vary by evaluation domain: (1) SimplerEnv: A single-view RGB image (256 × 256) is used across both stages; (2) LIBERO: Stage 1 uses a third-person view (200 × 200), supplemented by an additional wrist-view camera (200 × 200) for Stage 2; (3) Real-world: Stage 1 relies on the front-camera view (256 × 144), while Stage 2 incorporates a wrist-view image (240 × 135).

\noindent{\bf Structured planner training.} We train the world model to predict the next structured frame foresight from the instruction and visual context. We initialize the model with pretrained weights of world model from UniVLA~\cite{wang2025unifiedvisionlanguageactionmodel}. Supervision is applied only to vision tokens, guiding the model to generate physically grounded visual foresight. Because training solely on structured frames is temporally sparse, we apply a sliding-window augmentation that slightly shifts the target timestep and resamples its context, densifying supervision and exposing the model to richer spatiotemporal patterns. The structured planner is trained with a universal batch size of 32 for 30K, 10K, and 2K steps on the SimplerEnv, LIBERO, and real-world setups, respectively.

\noindent{\bf Action policy training.} We fine-tune the planner to shift from visual foresight prediction to action prediction. The input sequence begins with the language instruction, followed by interleaved visual and action tokens. Supervision is applied solely to action tokens, allowing the model to learn an explicit action policy grounded in visual planning. Subsequently, the action policy is trained across the three domains with varying configurations: 20K steps (batch size 128) for SimplerEnv, 4K steps (batch size 196) for LIBERO, and 2K steps (batch size 32) for real-world deployments.

 \noindent{\bf Efficient adaptation.}
To specialize the pre-trained world model for generating task-semantic structured frames without costly full-model retraining, we employ LoRA\cite{hu2022lora} (rank=32) for parameter-efficient fine-tuning. This approach preserves the model's foundational visual prediction capabilities while achieving high-quality foresight at a fraction of the computational cost, validating it as a highly effective adaptation strategy.

\subsection{Main Results}\label{sec:experimental_evaluation}

\noindent{\bf SimplerEnv benchmark} 
As shown in \cref{tab:simpler_env_widowx}, our StructVLA achieves an average success rate of 75.0\% on the SimplerEnv-WidowX benchmark, outperforming all prior baselines. Beyond the overall score, StructVLA achieves consistently strong performance across tasks. In particular, on the challenging \emph{Stack Green on Yellow Block} task, which requires sustained progress tracking and precise spatial execution, StructVLA matches VideoVLA and clearly outperforms the other baselines. We attribute these gains to structured frame foresight, which provides sparse, kinematically grounded progress anchors and helps stabilize closed-loop control over longer horizons.

\noindent{\bf LIBERO benchmark}
As shown in \cref{tab:libero_comparison}, our StructVLA achieves 94.8\% average success on LIBERO and shows strong performance overall, especially on LIBERO-Object, LIBERO-Goal, and LIBERO-Long. In particular, on the challenging LIBERO-Long suite, StructVLA performs notably better than prior methods, suggesting more reliable execution over extended horizons. We attribute this to the structured planner, which predicts structured frames at key interaction transitions and provides sparse mid-level visual subgoals for control. These subgoals act as progress anchors that guide the policy, reduce drift from accumulated errors, and improve consistency in multi-stage manipulation.

\begin{figure*}[t]
  \centering
  \includegraphics[width=0.95\textwidth]{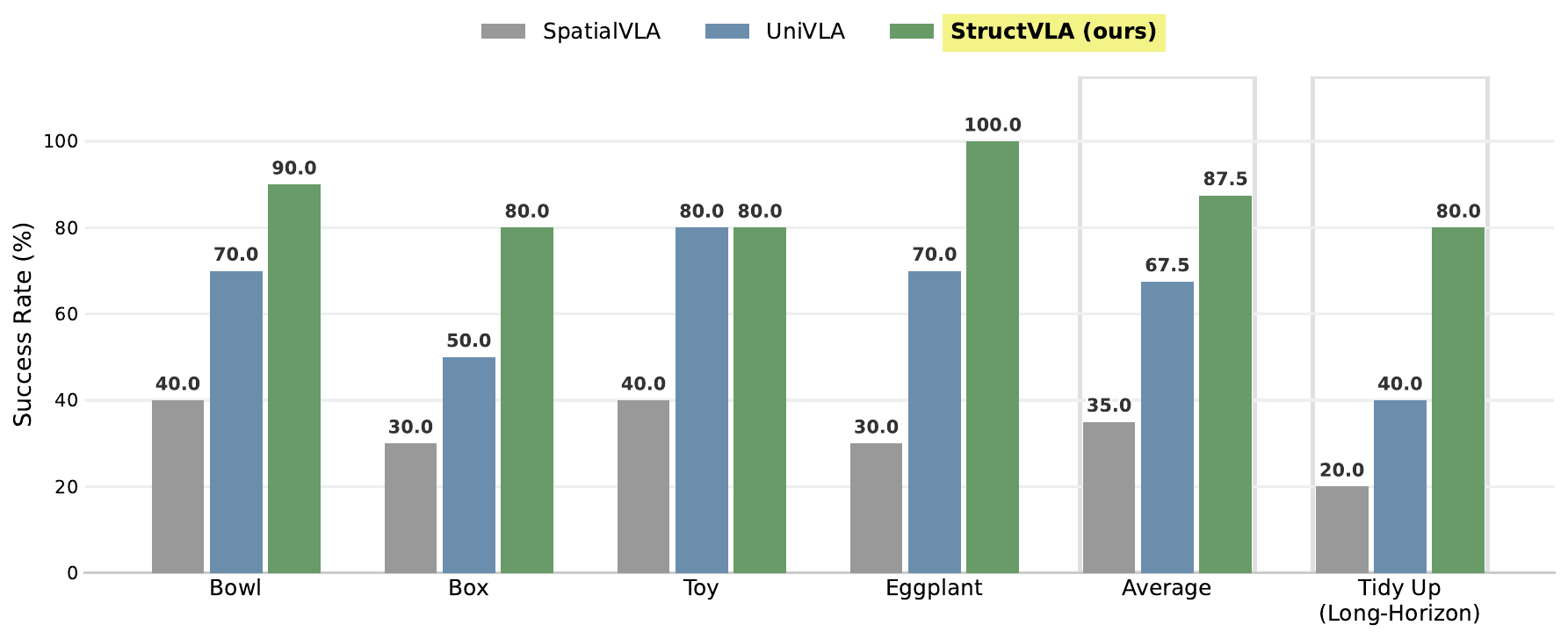}
  \vspace{-3mm}
\caption{
    \textbf{Quantitative Results on Real-World Deployments.} Success rates on diverse physical manipulation tasks (10 trials per task). \textbf{StructVLA} matches or exceeds prior baselines on the foundation tasks and remains strong on the challenging long-horizon tidy-up task.}
\label{fig:real_robot_summary}
\vspace{-3mm}
\end{figure*}

\noindent{\bf Real-world depolyments.} 
As shown in \cref{fig:real_robot_summary}, StructVLA outperforms the baselines in real-world deployments. On the four pick-and-place tasks, StructVLA achieves an 87.5\% average success rate, which is 20.0\% higher than UniVLA, and reaches 10/10 on the eggplant task. On the long-horizon \emph{tidy-up} task, StructVLA completes 8/10 trials, compared with 4/10 for UniVLA and 2/10 for SpatialVLA, indicating stronger stability over extended execution. These results suggest that incorporating structured planning improves real-world robustness, especially for long-horizon execution where small deviations and environmental noise can otherwise compound over time.

\begin{figure*}[t]
    \centering
    \begin{minipage}[b]{0.56\textwidth}
        \centering
        \includegraphics[width=\linewidth]{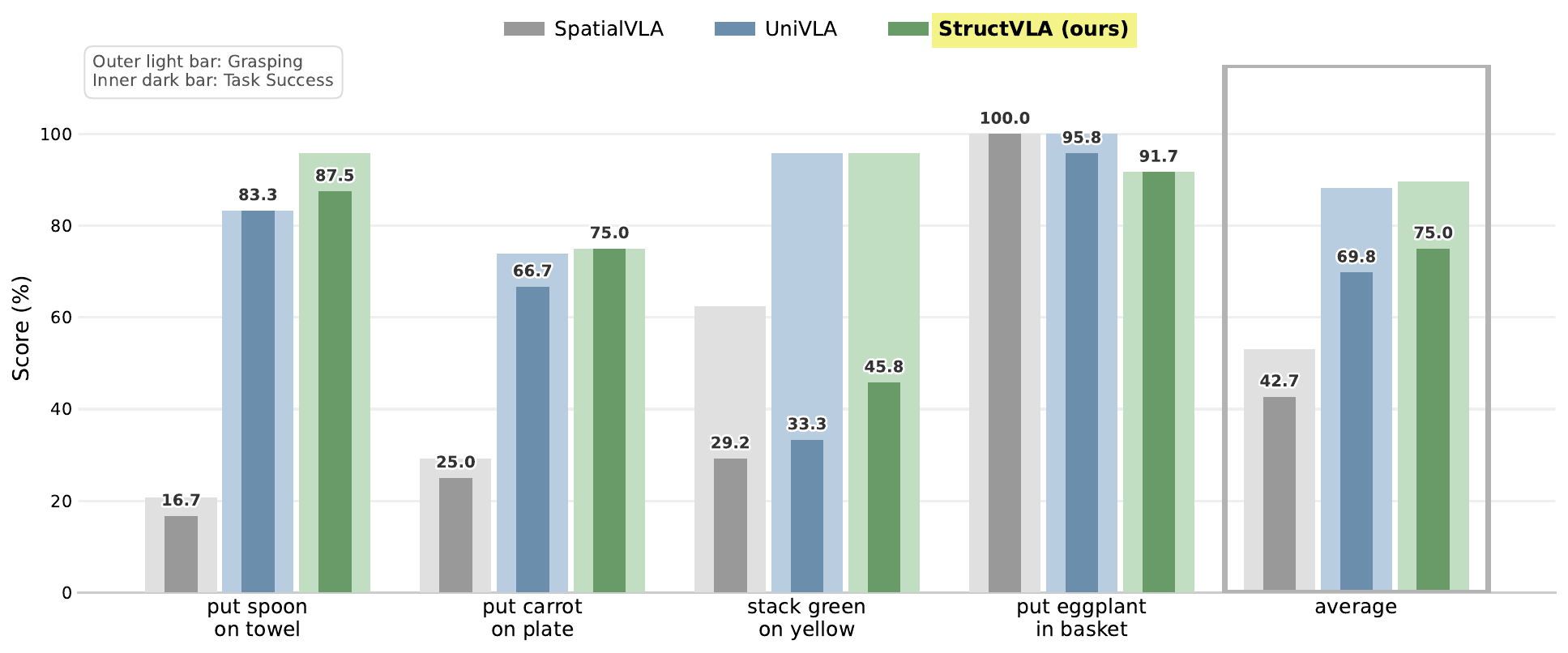}
        \caption{
        \textbf{Grasp Success vs. Task Success.} We analyze how grasp success translates to task completion. \textbf{StructVLA} attains the highest mean grasp success and a higher grasp-to-task conversion rate than the baseline UniVLA, indicating improved long-horizon stability with reduced drift and error accumulation.
        }
        \label{fig:performance_comparison}
    \end{minipage}\hfill 
    \begin{minipage}[b]{0.40\textwidth}
        \centering
        \includegraphics[width=\linewidth]{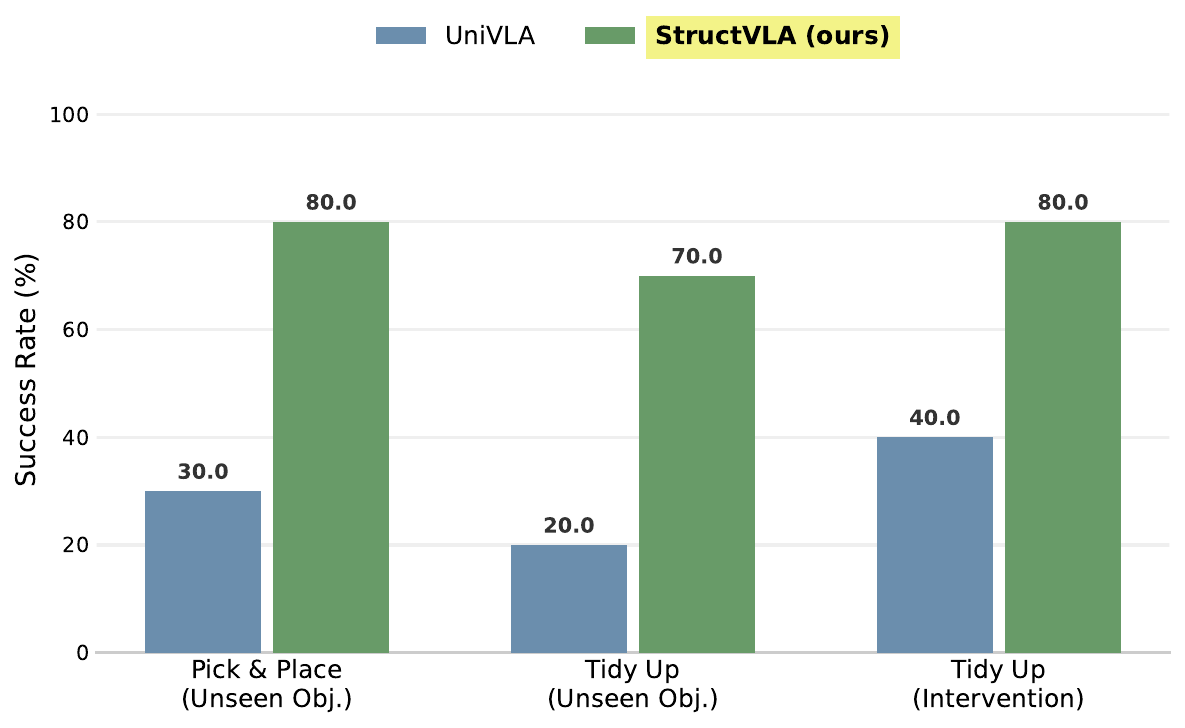}
        \caption{
        \textbf{Real-World Generalization and Robustness.} Success rates for zero-shot generalization (novel object geometries) and dynamic human interventions. \textbf{StructVLA} outperforms the baseline in all settings.
        }
        \label{fig:generalization}
    \end{minipage}
    \vspace{-5mm}
\end{figure*}

\noindent{\bf Real-world generalization and robustness.}
Beyond standard instruction following, we evaluate zero-shot generalization and robustness under dynamic perturbations (\cref{fig:generalization}). On out-of-distribution pick-and-place with unseen colors (\eg, a pink bowl) and novel geometries (\eg, a blue cup), StructVLA achieves 80.0\% success, compared to 30.0\% for UniVLA. On the long-horizon tidy-up task, StructVLA generalizes to unseen objects and maintains 70.0\% success. Under human interventions such as adding objects during execution, StructVLA retains 80.0\% success versus 40.0\% for UniVLA. We attribute these gains to structured planning with kinematically grounded milestones, which provides stable progress anchors and helps the policy remain consistent under appearance shifts and dynamic disturbances.

\begin{figure}[t]
  \centering
  \includegraphics[width=\columnwidth]{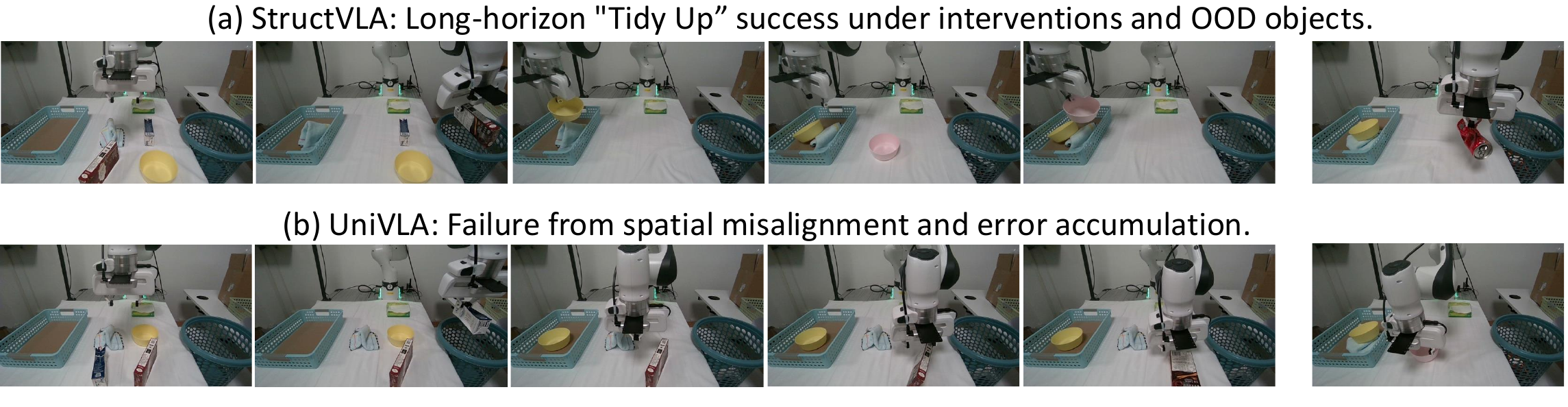}
\caption{
\textbf{Qualitative comparison of real-world deployments.} \textbf{(a) StructVLA (Ours):} Robust long-horizon execution with zero-shot generalization to OOD objects (\eg, pink bowl, Coke can). \textbf{(b) UniVLA (Baseline):} Prone to error accumulation and grasp failures, consistent with dense-prediction drift and limited generalization.
}

  \label{fig:real-results}
  \vspace{-2mm}
\end{figure}

\subsection{Extended Analysis}\label{sec: extend analysis}

Beyond the main performance results, this section provides a deeper analysis of StructVLA's core properties through a series of empirical investigations and ablation studies. Specifically, we first evaluate the performance and mechanistic insights of the proposed structured planner. We then examine the training efficiency of our approach and investigate the correlation between grasp quality and overall task success. Finally, we analyze the role of historical context in ensuring accurate predictions, and present an extended comparative evaluation of real-world deployments.

\noindent{\bf Structured planner performance. }
As shown in \cref{fig:planner_visualization}, our structured planner predicts the next structured frame from the current observation and instruction, maintaining coherent task progress over multi-step rollouts with less drift than a conventional world model. Even from a single observation, it rolls out a five-step sequence that stays consistent with the intended manipulation trajectory. In contrast, the baseline world model either produces near-static futures with weak progress cues or introduces distortions under larger state changes. These results suggest that training on structured frames yields more stable and informative foresight for long-horizon execution, providing intermediate visual targets that the downstream policy can follow more reliably.

\noindent{\bf Mechanistic insights of Structured planner. } 
We visualize attention maps from the final self-attention layer in \cref{fig:planner_more_visualization}(a) for both the structured planner and the baseline world model. During kinematic transitions, the planner places higher attention on motion related regions, most notably around the gripper, whereas the baseline exhibits a more diffuse distribution without a clear interaction focus. \cref{fig:planner_more_visualization}(b) evaluates the planner on out-of-distribution scenes and tasks excluded from Stage 1 training. The predictions remain visually coherent, suggesting that structured-frame finetuning preserves useful pretrained priors and generalizes to novel environments.

\noindent{\bf Training efficiency of StructVLA. }
As shown in \cref{tab:efficiency}, StructVLA exhibits strong sample efficiency, reaching high success rates with substantially fewer training steps than prior methods under the same batch size. We attribute this to StructVLA’s structured planning, which encourages representations that capture task progress and reduces the policy’s effective search space, improving learning efficiency. This reduces the learning burden on the action policy and enables faster improvement with less training data.

\noindent{\bf Analysis of Grasp Quality vs. Task Success.}
Beyond overall task success, \cref{fig:performance_comparison} examines how grasp success translates into task completion. StructVLA achieves the highest mean grasp success (89.6\%), although on \emph{Put Eggplant} it is slightly below two baselines (91.7\% vs.\ 100\%). More importantly, StructVLA attains a higher grasp-to-task conversion rate than the baseline UniVLA, indicating more consistent post-grasp execution. For instance, on \emph{Stack Green on Yellow Block}, StructVLA matches UniVLA in grasp success (95.8\%) but achieves higher task success (45.8\% vs.\ 33.3\%), suggesting that structured planning improves multi-step, long-horizon execution beyond grasping alone and may reduce plan drift and error accumulation.

\begin{table}[t]
  \centering
  
  \begin{minipage}[c]{0.48\linewidth}
    \centering
    \caption{
      \textbf{Learning Efficiency and Convergence Speed.} \textbf{Learning Efficiency and Convergence Speed.} Comparison of the training steps required for different models to reach peak success. All methods are trained under identical environments and hyperparameters (\eg, batch size 128). 
    }
    \label{tab:efficiency}
    \vspace{2mm} 
    \renewcommand{\arraystretch}{1.15}
    \begin{tabular}{@{}l@{\hspace{8pt}}c@{\hspace{8pt}}c@{\hspace{8pt}}c@{}}
      \toprule
      \textbf{Method} & \textbf{Batch} & \textbf{Steps} & \textbf{Accuracy} \\
      \midrule
      RoboVLMs & 128 & 50k & 37.5\% \\
      UniVLA & 128 & 12k & 64.8\% \\
      \textbf{StructVLA} & 128 & \textbf{10k} & \textbf{66.7\%} \\
      \bottomrule
    \end{tabular}
  \end{minipage}
  \hfill 
  \begin{minipage}[c]{0.48\linewidth}
    \centering
    \caption{
    \textbf{Historical Context Length.} Unlike UniVLA, StructVLA improves with longer context, effectively leveraging extended history for long-horizon reasoning and distant goal prediction.
    }
    \label{tab:history_window}
    \vspace{2mm} 
        \begin{tabular}{@{}l@{\hspace{9pt}}c@{\hspace{9pt}}l@{}}
          \toprule
          \textbf{Method} & \textbf{History} & \textbf{Accuracy} \\
          \midrule
            \multirow{3}{*}{StructVLA} & 2 & \makebox[3.6em][c]{70.8\%} \\
                                       & 3 & \makebox[3.6em][c]{71.9\%}\mhan{$_{\uparrow 1.55\%}$} \\
                                       & 4 & \makebox[3.6em][c]{\textbf{75.0\%}}\mhan{$_{\bf {\uparrow 5.93\%}}$} \\
            \midrule
            \multirow{3}{*}{UniVLA}    & 2 & \makebox[3.6em][c]{\textbf{69.8\%}} \\
                                       & 3 & \makebox[3.6em][c]{69.8\%}\blues{$_{-0.0}$} \\
                                       & 4 & \makebox[3.6em][c]{66.7\%}\blues{$_{\bf {\downarrow 4.44\%}}$} \\
          \bottomrule
        \end{tabular}
  \end{minipage}

\end{table}

\noindent{\bf History context for accurate prediction.}
On SimplerEnv, we evaluate history windows of 2, 3, and 4 in \cref{tab:history_window} and compare StructVLA with UniVLA under identical computation budgets. StructVLA improves as the history window increases, rising from 70.8\% to 75.0\%, whereas UniVLA drops from 69.8\% to 66.7\%. This ablation supports our claim that the structured planner benefits from longer temporal context because it uses the additional observations to forecast the next structured frame and capture long-range task progress, rather than being dominated by short-term visual details. As a result, larger history windows provide useful semantic and spatiotemporal evidence for planning and lead to better performance, instead of introducing distracting noise.

\noindent{\bf Comparative Evaluation of Real-World Deployments. }
As shown in \cref{fig:real-results}, StructVLA demonstrates reliable execution on the long-horizon \emph{tidy-up} task, successfully manipulating four objects and remaining stable under dynamic interventions with unseen objects such as a pink bowl and a coke can. In contrast, UniVLA shows weaker long-horizon stability and can suffer from error accumulation when task progress is not explicitly constrained. In our qualitative examples, UniVLA fails to grasp the towel and later knocks over the biscuit box due to compounding positional errors, and it also struggles to handle the novel pink bowl. Overall, these observations highlight a clear gap in deployment behavior, where StructVLA provides more consistent task completion and stronger zero-shot robust generalization in real-world settings.


%% file: sec/5_conclusion.tex
\section{Conclusion}
\label{sec:conclision}

We presented \textbf{StructVLA}, a novel Vision-Language-Action framework that reformulates generative world models into explicit, physically grounded structured planners for action control. To achieve robust long-horizon planning and strict physical alignment, StructVLA predicts discrete structured frames grounded in kinematic cues. Through a two-stage training paradigm within a unified token vocabulary, StructVLA bridges the gap between high-level planning signals and low-level action generation. Extensive empirical evaluations demonstrate that StructVLA achieves highly competitive performance, delivering a 75.0\% average success rate on the SimplerEnv-WidowX benchmark and 94.8\% on LIBERO. Real-world deployments further validate its exceptional cross-category generalization and robustness in executing complex, long-horizon tasks. Ultimately, StructVLA moves beyond dense video generation and semantic planning, providing physically grounded visual foresight for reliable robot control.

\subsubsection{Limitations and Discussion.}
While our results are promising, the primary limitation of this work is its experimental scale, constrained by computational resources. Future evaluation on larger, multi-domain datasets is thus essential to validate broader generalization. Promising research directions also include exploring advanced training strategies and architectural refinements to enhance model stability, and incorporating online reinforcement learning to improve the policy's robustness for real-world deployment.

%% file: sec/6_supplementation.tex
\setcounter{page}{1}
\renewcommand*{\sectionautorefname}{Sec.}
\renewcommand*{\subsectionautorefname}{Sec.}
\renewcommand*{\subsubsectionautorefname}{Sec.}
\renewcommand*{\paragraphautorefname}{Sec.}
\renewcommand*{\subparagraphautorefname}{Sec.}
\providecommand*{\Sectionautorefname}{Sec.}
\providecommand*{\Subsectionautorefname}{Sec.}
\providecommand*{\Subsubsectionautorefname}{Sec.}
\providecommand*{\Paragraphautorefname}{Sec.}
\providecommand*{\Subparagraphautorefname}{Sec.}
\setcounter{figure}{0}
\renewcommand{\thefigure}{S\arabic{figure}}
\setcounter{table}{0}
\renewcommand{\thetable}{S\arabic{table}}
\setcounter{equation}{0}
\renewcommand{\theequation}{S\arabic{equation}}
\section{Overview}

This supplementary material provides additional implementation details and further analyses for StructVLA. It is organized as follows.

In \cref{sec:implementation_details}, we present a detailed breakdown of the experimental setup to facilitate reproducibility. This includes the data curation pipeline and implementation details of the training setup.

In \cref{sec:visualizations}, we provide additional qualitative results and in-depth case studies. Through representative visualizations, we further illustrate StructVLA's planning and execution behavior across a diverse range of task scenarios.

We also provide a supplementary video demonstrating StructVLA on the real-world tidy-up task in the same folder as this supplementary PDF.

\section{Details}\label{sec:implementation_details}

In this section, we provide additional implementation details to facilitate reproducibility. We first describe the VLM-based offline filtering adopted during data curation, and then present the experimental setups and training configurations for the simulation benchmarks and real-world deployments..

\subsection{Details for VLM-based Offline Filtering}

After the initial candidate structured frames are extracted and saved in the keystep CSV file, we further refine them with a lightweight VLM-based filtering step. For each episode, we provide the VLM with the global task instruction, the candidate row indices from the original keystep CSV file, and the corresponding raw RGB images. Using the images as the primary evidence, the VLM removes only clearly unstable points, temporary false positives, and overly redundant candidates from dense local intervals. The retained row indices are then used to select the corresponding rows from the original CSV file and write a new filtered CSV file while preserving the original schema. In implementation, the VLM output is constrained to a fixed JSON schema containing a validity flag, an optional list of problems, and the retained row indices.

\paragraph{Prompt used for VLM-based offline filtering.}
\begin{tcolorbox}[
  colback=black!2,
  colframe=black!20,
  boxrule=0.3pt,
  arc=1mm,
  left=1mm,
  right=1mm,
  top=1mm,
  bottom=1mm,
  boxsep=1mm,
  breakable
]
\begin{Verbatim}[
  fontsize=\footnotesize,
  breaklines=true,
  breakanywhere=true,
  breaksymbolleft={},
  breaksymbolright={}
]
You are filtering candidate structured frames for a robot manipulation episode using images as the primary source of truth.

The input is an ordered list of candidate keysteps from the original CSV file, extracted from robot-centric kinematic cues. Each candidate is identified by its row index in the original CSV file, together with its image and the task instruction.

Your task is only to perform a conservative filtering of the candidates. Do not rewrite the instruction, do not add new events, and do not reconstruct the sequence.

Filtering goals:
1) Remove clearly unstable perturbation points caused by control noise, hardware jitter, or transient fluctuations.
2) Remove temporary false positives that clearly do not reflect a reasonable task progression.
3) Remove candidates that are visually too close to nearby neighbors and provide almost no additional information.

Important principles:
- Be conservative in filtering. Prefer keeping a candidate unless it is clearly abnormal or clearly redundant.
- When in doubt, keep the candidate rather than remove it.
- Preserve sufficient temporal coverage of the episode. The remaining candidates should still reflect the main progression of the manipulation process.
- Do not over-prune the sequence. In most normal cases, the filtered sequence should still contain multiple keyframes rather than collapsing to a single frame.
- Only remove a candidate when there is clear visual evidence that it is a noise point, a brief disturbance, or nearly duplicate with a nearby retained frame.
- If several nearby candidates correspond to essentially the same stable state, keep the earlier one.
- Preserve the original temporal order of the remaining candidates.
- Only delete candidates; do not add new ones.
- Use the images as the primary evidence.
- The task instruction is provided only as global context.
- Do not apply extra semantic rules beyond visible task coherence and local temporal consistency.

Return a JSON object with:
- filtered_ok: bool
- problems: [str, ...]
- kept_rows: [int, ...]
\end{Verbatim}
\end{tcolorbox}

\subsection{Implementation Details for Benchmarks}

\textbf{Structured planner.}
We train the structured planner by fine-tuning a pre-trained world model. Unless otherwise specified, we use AdamW with a peak learning rate of $2 \times 10^{-4}$, weight decay of $0.1$, and a cosine learning-rate schedule with warmup. At this stage, we primarily supervise the prediction of structured frames and disable gripper prediction, so that the planner focuses on forecasting future structured subgoals. Optionally, we retain a small auxiliary loss on the surrounding visual context to preserve the general visual modeling capability of the underlying world model. We adopt parameter-efficient fine-tuning via LoRA with rank $r = 32$ and scaling factor $\alpha = 32$, applied to the major attention and MLP projection layers, while keeping the backbone weights frozen. This allows the planner to adapt to the keystep-centric supervision without updating the full model.

\textit{SimplerEnv.}
We fine-tune the structured planner on the BridgeData V2 dataset using 8 H100 GPUs for approximately 40 hours. The model is trained for $30\mathrm{k}$ optimization steps. The per-device batch size is 2 with gradient accumulation of 2, resulting in an effective batch size of 32 sequences per optimization step. Each training sample contains up to 3 visual frames sampled around a structured frame with a temporal interval of 5 environment steps, providing local visual context for structured frame prediction, where we further apply a window-based sliding augmentation whose size matches the temporal interval to slightly shift the target timestep and resample its context.

\textit{LIBERO.}
We fine-tune the same pre-trained world model on a mixed dataset consisting of all four LIBERO suites, using 8 H100 GPUs for approximately 10 hours. The model is trained for $10\mathrm{k}$ optimization steps. The per-device batch size is 2 with gradient accumulation of 2, resulting in an effective batch size of 32 sequences per optimization step. Each training sample contains 4 visual frames with a temporal interval of 10 environment steps, where the same window-based sliding augmentation is applied with a size matched to the temporal interval.

\textbf{Action policy.}
In the second stage, we further fine-tune the structured planner from Stage 1 to learn an action policy. Unless otherwise specified, we use AdamW with a peak learning rate of $8 \times 10^{-5}$, weight decay of $0.1$, and a cosine learning-rate schedule with warmup. At this stage, we supervise action prediction using a FAST action tokenizer, so the model is optimized as an action policy conditioned on visual observations and language instructions.

\textit{SimplerEnv.}
We train the action policy using 8 H100 GPUs for approximately 30 hours. The model is trained for $20\mathrm{k}$ optimization steps. The per-device batch size is 4, and gradients are accumulated for 4 steps across 8 GPUs, yielding an effective batch size of 128 sequences per optimization step. Each training sample contains 2 visual frames and a horizon of 5 action steps arranged in an interleaved video format.

\textit{LIBERO.}
We train the action policy using 8 H100 GPUs for approximately 30 hours. The model is trained for $4\mathrm{k}$ optimization steps. The per-device batch size is 2, and gradients are accumulated for 12 steps across 8 GPUs, yielding an effective batch size of 192 sequences per optimization step. Each training sample contains 4 visual frames and a horizon of 10 action steps arranged in an interleaved video format. Unlike SimplerEnv, wrist-camera observations are included at this stage.

\textbf{Inference setting.}
At inference time, we evaluate the trained model in the SimplerEnv-WidowX environment under different history context lengths $L \in \{2,3,4\}$. Based on this ablation, we adopt $L = 4$ as the default setting and report the corresponding results as the final model performance. Unless explicitly stated otherwise, all quantitative results and analyses that do not specifically compare different history lengths are obtained with history context length $L = 4$. For LIBERO, we also use history context length $L = 4$, following the default setting selected based on the SimplerEnv history-length ablation.

\begin{figure*}[t]
  \centering
  \includegraphics[width=1.0\textwidth]{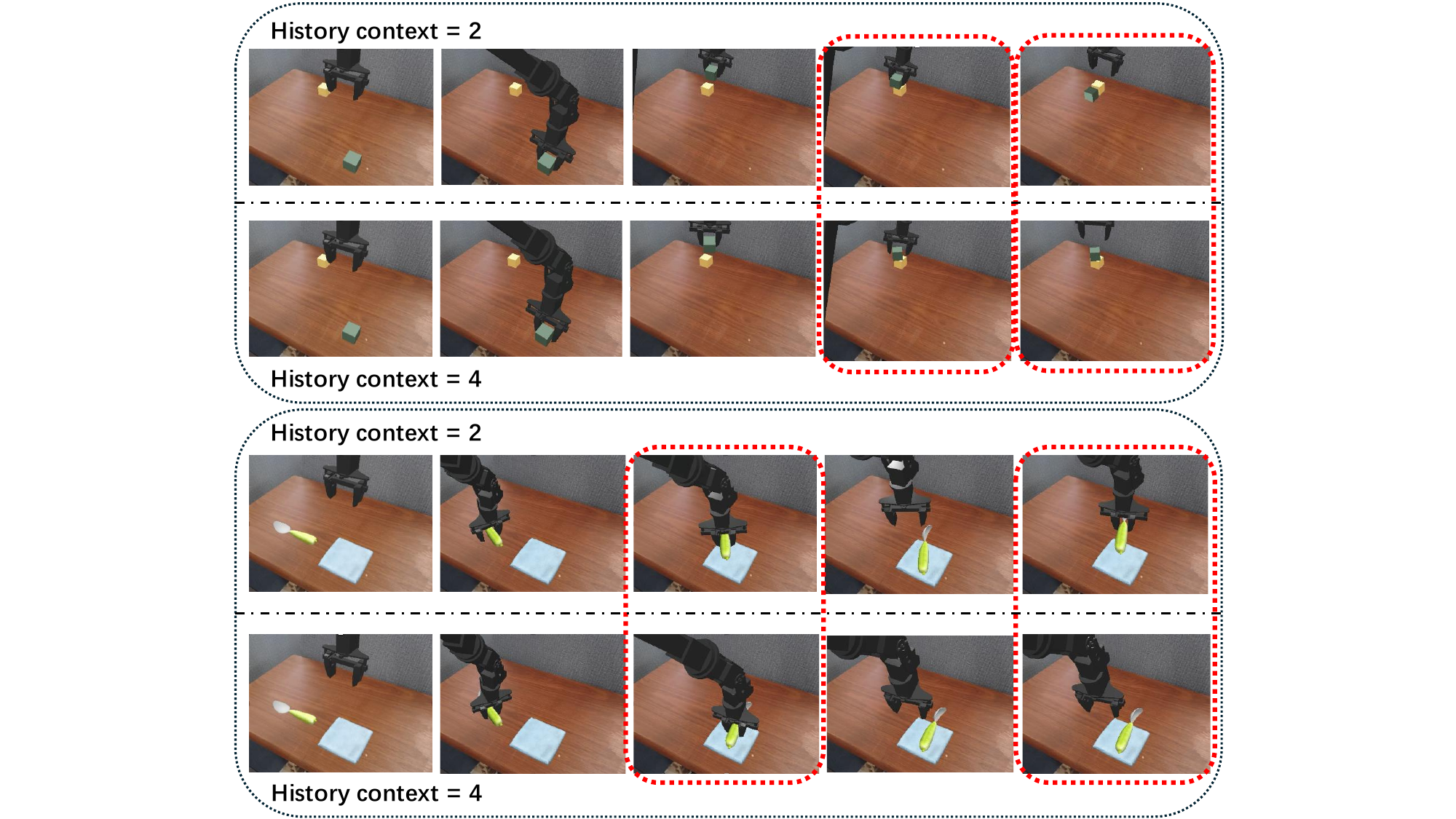}
\caption{\textbf{Qualitative analysis of history context length on task execution.}
We compare policy behavior under different history context lengths ($L$) on two representative tasks: ``Stack green cube on yellow cube'' (top) and ``Put spoon on towel'' (bottom). With a short history ($L=2$), the policy lacks sufficient spatio-temporal context to accurately infer task progress and interaction geometry. \textbf{Top:} this leads to imprecise placement, causing the cube to slip during stacking (red dashed box). \textbf{Bottom:} the policy exhibits temporal ambiguity and incorrectly re-grasps the object after it has already been placed. In contrast, a longer history ($L=4$) provides richer temporal evidence and motion context, enabling accurate execution in both cases.}
  \label{fig:history_figure}
\end{figure*}

\begin{figure*}[t]
  \centering
  \includegraphics[width=1.0\textwidth]{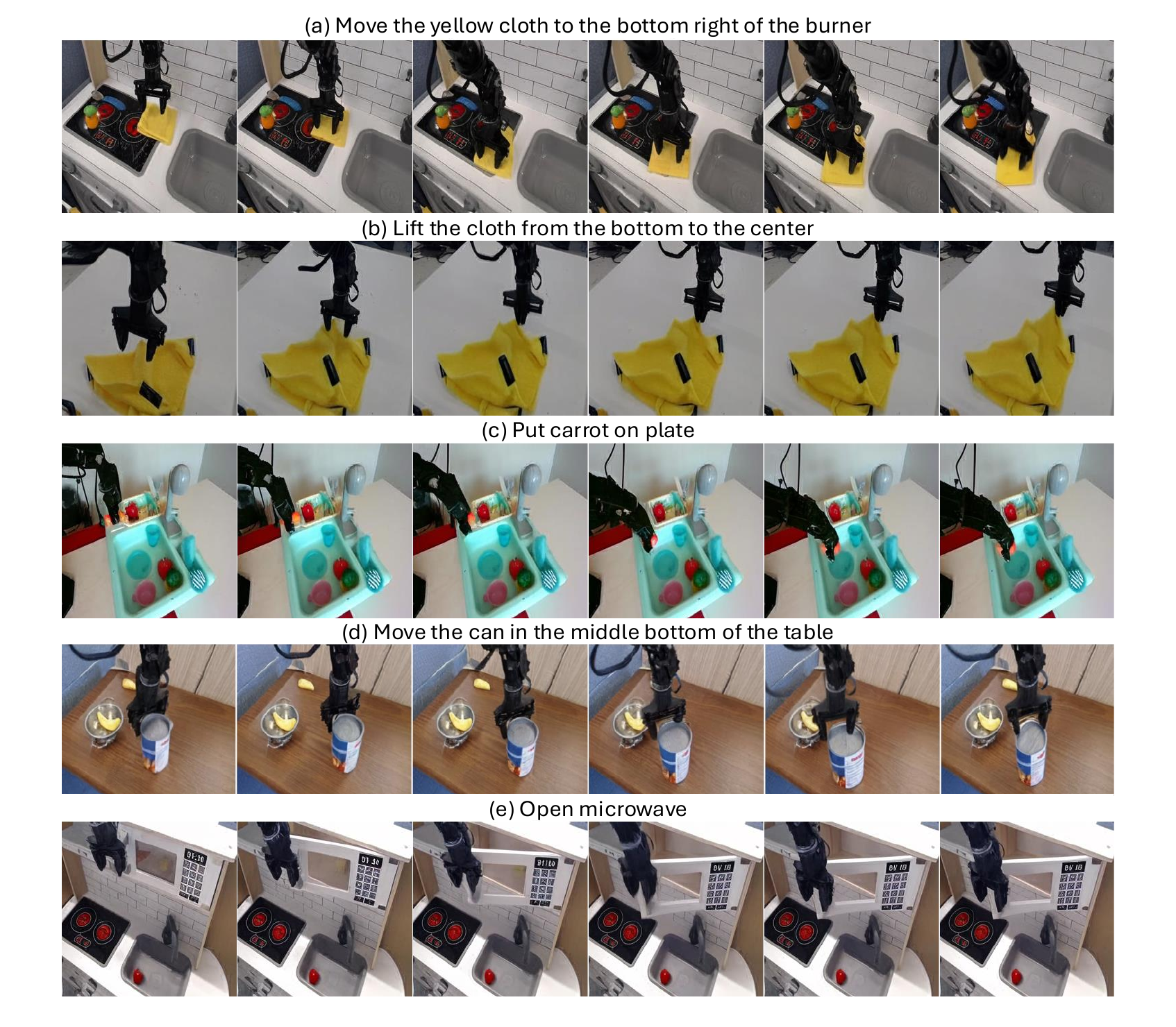}
\caption{\textbf{Extended visual foresight predictions of the structured planner.}
These examples show that our planner predicts sparse yet informative visual subgoals that remain temporally coherent and physically grounded across diverse manipulation tasks. The predicted structured frames align with key interaction stages, such as approach, contact, transfer, and placement, supporting the central design of StructVLA: replacing dense future rollouts with compact structured foresight for control. In (e), we show a more challenging case involving complex kinematics, particularly for articulated or highly constrained object motions.}
  \label{fig:more_visual_results}
\end{figure*}
\subsection{Implementation Details on Real-world Deployment
}
As described in the main paper, we deploy our method on a real Franka robot arm and conduct experiments on both basic pick-and-place tasks and long-horizon tidy-up tasks. We conduct two separate training runs corresponding to these task categories, and the detailed training procedures are described below.

\textit{Structured planner.}
For real-world experiments, we fine-tune the same pre-trained world model on a real-robot dataset using 4 H100 GPUs for approximately 3 hours in the first-stage structured-planner training. We use AdamW with a peak learning rate of $6 \times 10^{-5}$, weight decay of $0.1$, and a cosine learning-rate schedule with warmup for $2\mathrm{k}$ optimization steps. The per-device batch size is 4 with gradient accumulation of 2, resulting in an effective batch size of 32 sequences per optimization step. Each training sample contains 4 visual frames from the front-view camera only. The temporal interval is set to 3 steps for pick-and-place tasks and 4 steps for tidy-up tasks, while the window-based sliding augmentation size is fixed to 3 in both settings. At this stage, we supervise only visual prediction. Unlike the simulated benchmarks, we directly fine-tune the model on the limited real-robot data with full-parameter training rather than using LoRA-based parameter-efficient fine-tuning.

\textit{Action policy.}
We further fine-tune the structured planner from Stage 1 to learn the action policy. Training is performed on 4 H100 GPUs for approximately 3 hours. We use AdamW with a peak learning rate of $6 \times 10^{-5}$, weight decay of $0.1$, and a cosine learning-rate schedule with warmup for $2\mathrm{k}$ optimization steps. The per-device batch size is 4 with gradient accumulation of 2, resulting in an effective batch size of 32 sequences per optimization step. Each training sample contains 2 visual frames and a horizon of 5 action steps arranged in an interleaved video format. At this stage, we supervise only action prediction using a FAST action tokenizer, and wrist-camera observations are also included.

\textit{Inference setting.} For real-world deployment, we adopt a server-client inference architecture. The trained model is hosted on a remote RTX 4090 GPU server and exposed through a Flask-based API. During execution, the robot-side client sends the current task description together with synchronized observations from the front and wrist cameras to the server. The server receives the incoming images, performs model inference, and returns a predicted action chunk to the robot client for execution. Empirically, we execute the first two actions from each predicted action chunk at every control cycle. This design decouples model inference from on-robot control and enables efficient interaction between the robot platform and the deployed policy model.

For the baselines, we follow training conditions that are matched as closely as possible to those of our method, including the data setting and optimization protocol where applicable, to ensure fair comparison.

\section{In-depth analysis and visualizations}\label{sec:visualizations}
In this section, we provide additional qualitative analyses to complement the quantitative results. Through a set of representative visualizations, we further examine the behavior of StructVLA and highlight how the structured planner captures task progress and supports long-horizon control.

\subsection{Task Understanding and History Context}

We further present qualitative rollouts from the SimplerEnv environment to analyze how history context length affects task understanding and execution. When the available history is short, the planner may fail to accurately infer the current interaction stage and task progress, which can lead to suboptimal decisions. In some cases, this manifests as incorrect intermediate actions; in others, the policy continues issuing commands even after the task has effectively been completed. In StructVLA, increasing the history window substantially alleviates these issues, since richer temporal evidence allows the structured planner to better disambiguate the current state and form more stable long-horizon plans.

As illustrated in the bottom example of \cref{fig:history_figure}, under short-history conditions StructVLA may misinterpret task status and occasionally attempt to re-execute an action that has already been completed. This behavior is consistent with the design of our model: StructVLA is built upon long-horizon visual foresight and predicts structured subgoals over extended spatiotemporal horizons, so sufficient temporal context is important for accurately grounding the current task phase. At the same time, compared with models based on dense future prediction, StructVLA reasons over sparse structured planning targets and can therefore make more effective use of extended context for planning and prediction, while being less affected by redundant visual information.

\subsection{Action Accuracy and Window-based Sliding Augmentation}

During structured planner training, we adopt a window-based sliding augmentation strategy, in which the target timestep is slightly shifted and its local context is resampled around the annotated structured frame, thereby densifying supervision and exposing the planner to richer spatiotemporal patterns. This allows the model to observe a broader range of task phases and encourages it to model short-horizon future progression even when the current observation is already close to a structured frame. In practice, this improves execution efficiency and helps prevent the planner from becoming overly conservative around key stages of the trajectory. At the same time, the augmentation may introduce a mild temporal offset, in the sense that the predicted structured frame can be slightly shifted relative to the ideal next-step target, which may occasionally influence fine-grained execution precision.

Our qualitative analysis in \cref{fig:history_figure} suggests that this temporal offset can sometimes appear as minor execution bias when historical context is limited. For example, in the top example, it is reflected as slight spatial imprecision during placement. Increasing the amount of historical context helps alleviate this effect, since richer temporal evidence allows the planner to better resolve the intended interaction stage and compensate for local prediction offsets. This may be further explored through training or architectural refinements that improve fine-grained temporal alignment while preserving the robustness benefits of augmented supervision.

\subsection{More Visual Predictions from Structured Planner}

\cref{fig:more_visual_results} presents a broader set of representative predictions generated by our structured planner. These visualizations further demonstrate that the planner maintains strong long-horizon semantic understanding and spatiotemporal reasoning across diverse manipulation scenarios. Notably, consistent with the examples shown in \cref{fig:more_visual_results}, the planner still produces temporally coherent and task-relevant visual foresight under the window-based supervision strategy, indicating that this training design does not compromise the overall quality of structured planning.

However, we observe a more noticeable challenge in articulated manipulation tasks, such as the ``Open microwave'' scenario shown in \cref{fig:more_visual_results}(e). Unlike rigid pick-and-place tasks, the key progress transitions in hinge-driven interactions are not naturally aligned with discrete gripper-state changes, but instead unfold through continuous geometric motion. As a result, our current structured-frame extraction strategy, which is primarily built on robot-centric kinematic cues and contact-related transitions, is less well matched to these articulated motion patterns. While the planner can still capture the high-level goal progression, \eg, the microwave door being opened, the corresponding fine-grained arm motion is less precise. This limitation suggests that future improvements should focus on more geometry-aware structured-frame extraction for articulated tasks, for example by incorporating continuous motion-phase cues such as relative object pose evolution or articulation-state changes, so that the planner can better align visual foresight with the underlying interaction dynamics.